\documentclass{article}

\usepackage{times}
\usepackage{latexsym}
\usepackage{url}
\usepackage{graphicx}
\graphicspath{ {figures/} }
\usepackage{subcaption}
\usepackage{enumitem}
\usepackage{amsmath}
\usepackage{amssymb}
\usepackage[utf8]{inputenc}
\usepackage[vietnamese=nohyphenation]{hyphsubst}
\usepackage[vietnamese, english]{babel}
\usepackage{kotex}

\newcommand{\VI}{\foreignlanguage{vietnamese}}
\usepackage{multirow}

\usepackage{hyperref}

\usepackage[accepted]{icml2019}

\icmltitlerunning{Language Identification on Massive Datasets of Short Message using an Attention Mechanism CNN}

\begin{document}

\twocolumn[
\icmltitle{Language Identification on Massive Datasets of Short Message\\ using an Attention Mechanism CNN}



\icmlsetsymbol{equal}{*}

\begin{icmlauthorlist}
\icmlauthor{Duy Tin Vo}{equal,ct}
\icmlauthor{Richard Khoury}{equal,lv}
\end{icmlauthorlist}

\icmlaffiliation{ct}{Chata.ai, Calgary, Alberta, Canada}
\icmlaffiliation{lv}{Department of Computer Science and Software Engineering, Université Laval, Quebec City, Quebec, Canada}

\icmlcorrespondingauthor{Duy Tin Vo}{duytinvo@gmail.com}
\icmlcorrespondingauthor{Richard Khoury}{Richard.Khoury@ift.ulaval.ca}

\icmlkeywords{Deep Learning, Language Identification, Convolution Neural Network, Attention Mechanism, Twitter}

\vskip 0.3in
]




\begin{abstract}

Language Identification (LID) is a challenging task, especially when the input texts are short and noisy such as posts and statuses on social media or chat logs on gaming forums. The task has been tackled by either designing a feature set for a traditional classifier (e.g. Naive Bayes) or applying a deep neural network classifier (e.g. Bi-directional Gated Recurrent Unit, Encoder-Decoder). These methods are usually trained and tested on a huge amount of private data, then used and evaluated as off-the-shelf packages by other researchers using their own datasets, and consequently the various results published are not directly comparable. In this paper, we first create a new massive labelled dataset based on one year of Twitter data. We use this dataset to test several existing language identification systems, in order to obtain a set of coherent benchmarks, and we make our dataset publicly available so that others can add to this set of benchmarks. Finally, we propose a shallow but efficient neural LID system, which is a ngram-regional convolution neural network enhanced with an attention mechanism. Experimental results show that our architecture is able to predict tens of thousands of samples per second and surpasses all state-of-the-art systems with an improvement of 5\%.
\end{abstract}


\section{Introduction}\label{intro}
Language Identification (LID) is the Natural Language Processing (NLP) task of automatically recognizing the language that a document is written in. While this task was called "solved" by some authors over a decade ago, it has seen a resurgence in recent years thanks to the rise in popularity of social media \cite{jauhiainen2018automatic,jaech2016hierarchical}, and the corresponding daily creation of millions of new messages in dozens of different languages including rare ones that are not often included in language identification systems. Moreover, these messages are typically very short (Twitter messages were until recently limited to 140 characters) and very noisy (including an abundance of spelling mistakes, non-word tokens like URLs, emoticons, or hashtags, as well as foreign-language words in messages of another language), whereas LID was solved using long and clean documents. Indeed, several studies have shown that LID systems trained to a high accuracy on traditional documents suffer significant drops in accuracy when applied to short social-media texts \cite{lui2012langid,carter2013microblog}.

Given its massive scale, multilingual nature, and popularity, Twitter has naturally attracted the attention of the LID research community. Several attempts have been made to construct LID datasets from that resource. However, a major challenge is to assign each tweet in the dataset to the correct language among the more than 70 languages used on the platform. The three commonly-used approaches are to rely on human labeling \cite{lui2014accurate,tromp2011graph}, machine detection \cite{tromp2011graph,jurgens2017incorporating}, or user geolocation \cite{carter2013microblog,blodgett2017dataset,bergsma2012language}. Human labeling is an expensive process in terms of workload, and it is thus infeasible to apply it to create a massive dataset and get the full benefit of Twitter's scale. Automated LID labeling of this data creates a noisy and imperfect dataset, which is to be expected since the purpose of these datasets is to create new and better LID algorithms. And user geolocation is based on the assumption that users in a geographic region use the language of that region; an assumption that is not always correct, which is why this technique is usually paired with one of the other two. Our first contribution in this paper is to propose a new approach to build and automatically label a Twitter LID dataset, and to show that it scales up well by building a dataset of over 18 million labeled tweets. Our hope is that our new Twitter dataset will become a benchmarking standard in the LID literature.

\begin{table*}
\caption{Summary of literature results}
\begin{tabular}{lllcrl}
\hline
 Paper & Input & Algorithm & Metric & Results & langid.py \\
 \hline
 Tromp \& Pechenizkiy \yrcite{tromp2011graph} & Character n-grams & Graph & Accuracy & 0.975 &  0.941 \\
 Carter et al. \yrcite{carter2013microblog} & Social network information & Prior probabilities & Accuracy & 0.972 & 0.886\\
 Gamallo et al. \yrcite{gamallo2014comparing} & Words & Dictionary & F1-score & 0.733 & N/A\\
 Jaech et al. \yrcite{jaech2016hierarchical} & Words & LSTM & F1-score & 0.912 & 0.879\\
 Kocmi \& Bojar \yrcite{kocmi2017lanidenn}  & Character n-grams& GRU & Accuracy & 0.955 & 0.912\\
 Jurgens et al. \yrcite{jurgens2017incorporating}  & Character n-grams & Encoder-decoder & Accuracy & 0.982 & 0.960\\ \hline
 \end{tabular}
\label{table:background}
\end{table*}

Traditional LID models \cite{lui2012langid,carter2013microblog,gamallo2014comparing} proposed different ideas to design a set of useful features. This set of features is then passed to traditional machine learning algorithms such as Naive Bayes (NB). The resulting systems are capable of labeling thousands of inputs per second with moderate accuracy. Meanwhile, neural network models \cite{kocmi2017lanidenn,jurgens2017incorporating} approach the problem by designing a deep and complex architecture like gated recurrent unit (GRU) or encoder-decoder net. These models use the message text itself as input using a sequence of character embeddings, and automatically learn its hidden structure via a deep neural network. Consequently, they obtain better results in the task but with an efficiency trade-off. To alleviate these drawbacks, our second contribution in this paper is to propose a shallow but efficient neural LID algorithm. We followed previous neural LID \cite{kocmi2017lanidenn,jurgens2017incorporating} in using character embeddings as inputs. However, instead of using a deep neural net, we propose to use a shallow ngram-regional convolution neural network (CNN) with an attention mechanism to learn input representation. We experimentally prove that the ngram-regional CNN is the best choice to tackle the bottleneck problem in neural LID. We also illustrate the behaviour of the attention structure in focusing on the most important features in the text for the task. Compared with other benchmarks on our Twitter datasets, our proposed model consistently achieves new state-of-the-art results with an improvement of 5\% in accuracy and F1 score and a competitive inference time.      

The rest of this paper is structured as follows. After a background review in the next section, we will present our Twitter dataset in Section \ref{TwitterDataset}. Our novel LID algorithm will be the topic of Section \ref{ourwork}. We will then present and analyze some experiments we conducted with our algorithm in Section \ref{results}, along with benchmarking tests of popular and literature LID systems, before drawing some concluding remarks in Section \ref{conclusion}. Our Twitter dataset and our LID algorithm's source code are publicly available\footnote{\url{https://github.com/duytinvo/LID\_NN}}.

\section{Related Work}\label{RelatedWork}
In this section, we will consider recent advances on the specific challenge of language identification in short text messages. Readers interested in a general overview of the area of LID, including older work and other challenges in the area, are encouraged to read the thorough survey of \cite{jauhiainen2018automatic}.

\subsection{Probabilistic LID}
One of the first, if not the first, systems for LID specialized for short text messages is the graph-based method of \cite{tromp2011graph}. Their graph is composed of vertices, or character n-grams (n = 3) observed in messages in all languages, and of edges, or connections between successive n-grams weighted by the observed frequency of that connection in each language. Identifying the language of a new message is then done by identifying the most probable path in the graph that generates that message. Their method achieves an accuracy of 0.975 on their own Twitter corpus.\\
Carter, Weerkamp, and Tsagkias proposed an approach for LID that exploits the very nature of social media text \cite{carter2013microblog}. Their approach computes the prior probability of the message being in a given language independently of the content of the message itself, in five different ways: by identifying the language of external content linked to by the message, the language of previous messages by the same user, the language used by other users explicitly mentioned in the message, the language of previous messages in the on-going conversation, and the language of other messages that share the same hashtags. They achieve a top accuracy of 0.972 when combining these five priors with a linear interpolation.\\
One of the most popular language identification packages is the langid.py library proposed in \cite{lui2012langid}, thanks to the fact it is an open-source, ready-to-use library written in the Python programming language. It is a multinomial Naïve Bayes classifier trained on character n-grams (1 $\leq$ n $\leq$ 4) from 97 different languages. The training data comes from longer document sources, both formal ones (government publications, software documentation, and newswire) and informal ones (online encyclopedia articles and websites). While their system is not specialized for short messages, the authors claim their algorithm can generalize across domains off-the-shelf, and they conducted experiments using the Twitter datasets of \cite{tromp2011graph} and \cite{carter2013microblog} that achieved accuracies of 0.941 and 0.886 respectively, which is weaker than the specialized short-message LID systems of \cite{tromp2011graph} and \cite{carter2013microblog}.\\
Starting from the basic observation of Zipf's Law, that each language has a small number of words that occur very frequently in most documents, the authors of \cite{gamallo2014comparing} created a dictionary-based algorithm they called Quelingua. This algorithm includes ranked dictionaries of the 1,000 most popular words of each language it is trained to recognize. Given a new message, recognized words are given a weight based on their rank in each language, and the identified language is the one with the highest sum of word weights. Quelingua achieves an F1-score of 0.733 on the TweetLID competition corpus \cite{zubiaga2014overview}, a narrow improvement over a trigram Naïve Bayes classifier which achieves an F1-Score of 0.727 on the same corpus, but below the best results achieved in the competition.

\subsection{Neural Network LID}
Neural network models have been applied on many NLP problems in recent years with great success, achieving excellent performance on challenges ranging from text classification \cite{Vo18} to sequence labeling \cite{yang2018ncrf}. 
In LID, the authors of \cite{jaech2016hierarchical} built a hierarchical system of two neural networks. The first level is a Convolutional Neural Network (CNN) that converts white-space-delimited words into a word vector. The second level is a Long-Short-Term Memory (LSTM) network (a type of recurrent neural network (RNN)) that takes in sequences of word vectors outputted by the first level and maps them to language labels. They trained and tested their network on Twitter's official Twitter70 dataset\footnote{\url{https://blog.twitter.com/engineering/en_us/a/2015/evaluating-language-identification-performance.html}}, and achieved an F-score of 0.912, compared to langid.py's performance of 0.879 on the same dataset. They also trained and tested their system using the TweetLID corpus and achieved an F1-score of 0.762, above the system of \cite{gamallo2014comparing} presented earlier, and above the top system of the TweetLID competition, the SVM LID system of \cite{hurtado2014elirf} which achieved an F1-score of 0.752.\\
The authors of \cite{kocmi2017lanidenn} also used a RNN system, but preferred the Gated Recurrent Unit (GRU) architecture to the LSTM, indicating it performed slightly better in their experiments. Their system breaks the text into non-overlapping 200-character segments, and feeds character n-grams (n = 8) into the GRU network to classify each letter into a probable language. The segment's language is simply the most probable language over all letters, and the text's language is the most probable language over all segments. The authors tested their system on short messages, but not on tweets; they built their own corpus of short messages by dividing their data into 200-character segments. On that corpus, they achieve an accuracy of 0.955, while langid.py achieves 0.912. \\
The authors of \cite{jurgens2017incorporating} also created a character-level LID network using a GRU architecture, in the form of a three-layer encoder-decoder RNN. They trained and tested their system using their own Twitter dataset, and achieved an F1-score of 0.982, while langid.py achieved 0.960 on the same dataset.\\
To summarize, we present the key results of the papers reviewed in this section in Table \ref{table:background}, along with the results langid.py obtained on the same datasets as benchmark. 

\begin{table}
\caption{Twitter corpus distribution by language label.}
\tabcolsep=0.15cm
\begin{tabular}{lr|lr|lr}
\hline
Lang & P(\%) & Lang & P(\%) & Lang & P(\%) \\
\hline
EN & 36.458 & EL & 0.086 & LV, BG, \\
JA & 23.750 & SV & 0.046 & ~~~UR, TA & $10^{-3}$\\
ES & 9.964 & FA & 0.027 & MR, BN, \\
AR & 7.627 & VI & 0.021 & ~~~MR, BN,\\
PT & 6.839 & FI & 0.020 & ~~~IN, KN,\\
KO & 5.559 & CS & 0.015 & ~~~ET, SL,\\
TH & 2.965 & UK & 0.015 & ~~~GU, CY,\\
FR & 2.180 & HI & 0.013 & ~~~ZH, CKB,\\
TR & 2.152 & DA & 0.007 & ~~~IS, LT,\\
RU & 0.948 & HU & 0.006 & ~~~ML, SI,\\
IT & 0.490 & NO & 0.005 & ~~~IW, NE,\\
DE & 0.356 & RO & 0.003 & ~~~KM, MY,\\
PL & 0.251 & SR & 0.003 & ~~~TL, KA,\\
NL & 0.187 & EU & 0.002 & ~~~BO & $<10^{-3}$ \\
\hline
\end{tabular}
\label{table:datasetcompose}
\end{table}

\section{Our Twitter LID Datasets} \label{TwitterDataset}
\subsection{Source Data and Language Labeling}
Unlike other authors who built Twitter datasets, we chose not to mine tweets from Twitter directly through their API, but instead use tweets that have already been downloaded and archived on the Internet Archive\footnote{\url{https://archive.org/details/twitterstream}}. This has two important benefits: this site makes its content freely available for research purposes, unlike Twitter which comes with restrictions (especially on distribution), and the tweets are backed-up permanently, as opposed to Twitter where tweets may be deleted at any time and become unavailable for future research or replication of past studies. The Internet Archive has made available a set of 1.7 billion tweets collected over the year of 2017 in a 600GB JSON file which includes all tweet metadata attributes\footnote{\url{https://developer.twitter.com/en/docs/tweets/data-dictionary/overview/intro-to-tweet-json.html}}. Five of these attributes are of particular importance to us. They are $\it{tweet.id}$, $\it{tweet.user.id}$, $\it{tweet.text}$, $\it{tweet.lang}$, and $\it{tweet.user.lang}$, corresponding respectively to the unique tweet ID number, the unique user ID number, the text content of the tweet in UTF-8 characters, the tweet's language as determined by Twitter's automated LID software, and the user's self-declared language. \\
We begin by filtering the corpus to keep only those tweets where the user's self-declared language and the tweet's detected language correspond; that language becomes the tweet's correct language label. This operation cuts out roughly half the tweets, and leaves us with a corpus of about 900 million tweets in 54 different languages. Table \ref{table:datasetcompose} shows the distribution of languages in that corpus. Unsurprisingly, it is a very imbalanced distribution of languages, with English and Japanese together accounting for 60\% of all tweets. This is consistent with other studies and statistics of language use on Twitter, going as far back as 2013\footnote{https://www.statista.com/statistics/267129/most-used-languages-on-twitter/}. It does however make it very difficult to use this corpus to train a LID system for other languages, especially for one of the dozens of seldom-used languages. This was our motivation for creating a balanced Twitter dataset. 


\subsection{Our Balanced Datasets}
When creating a balanced Twitter LID dataset, we face a design question: should our dataset seek to maximize the number of languages present, to make it more interesting and challenging for the task of LID, but at the cost of having fewer tweets per language to include seldom-used languages. Or should we maximize the number of tweets per language to make the dataset more useful for training deep neural networks, but at the cost of having fewer languages present and eliminating the seldom-used languages. To circumvent this issue, we propose to build three datasets: a small-scale one with more languages but fewer tweets, a large-scale one with more tweets but fewer languages, and a medium-scale one that is a compromise between the two extremes. Moreover, since we plan for our datasets to become standard benchmarking tools, we have subdivided the tweets of each language in each dataset into training, validation, and testing sets.
\begin{itemize}
\item \textbf{Small-scale dataset:} This dataset is composed of 28 languages with 13,000 tweets per language, subdivided into 7,000 training set tweets, 3,000 validation set tweets, and 3,000 testing set tweets. There is thus a total of 364,000 tweets in this dataset. Referring to Table \ref{table:datasetcompose}, this dataset includes every language that represents 0.002\% or more of the Twitter corpus. To be sure, it is possible to create a smaller dataset with all 54 languages but much fewer tweets per language, but we feel that this is the lower limit to be useful for training LID deep neural systems.
\item \textbf{Medium scale dataset:} This dataset keeps 22 of the 28 languages of the small-scale dataset, but has 10 times as many tweets per language. In other words, each language has a 70,000-tweet training set, a 30,000-tweet validation set, and a 30,000-tweet testing set, for a total of 2,860,000 tweets.
\item \textbf{Large-scale dataset:} Once again, we increased tenfold the number of tweets per language, and kept the 14 languages that had sufficient tweets in our initial 900 million tweet corpus. This gives us a dataset where each language has 700,000 tweets in its training set, 300,000 tweets in its validation set, and 300,000 tweets in its testing set, for a total 18,200,000 tweets. Referring to Table \ref{table:datasetcompose}, this dataset includes every language that represents 0.1\% or more of the Twitter corpus.
\end{itemize}
\begin{figure*}[tbp]
\centering
\begin{subfigure}[t]{.3\textwidth}
    \centering
	\includegraphics[width=0.9\linewidth]{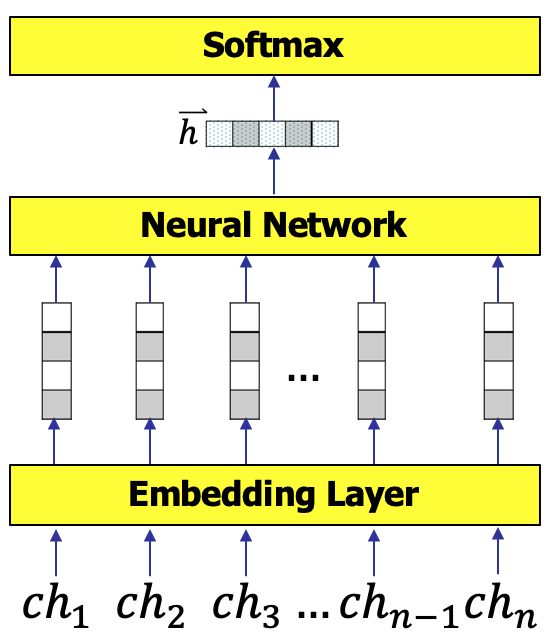}
    \caption{Neural network LID baseline.}
	\label{fig:nnbaseline}
\end{subfigure}
~
\begin{subfigure}[t]{.65\textwidth}
    \centering
	\includegraphics[width=0.95\linewidth]{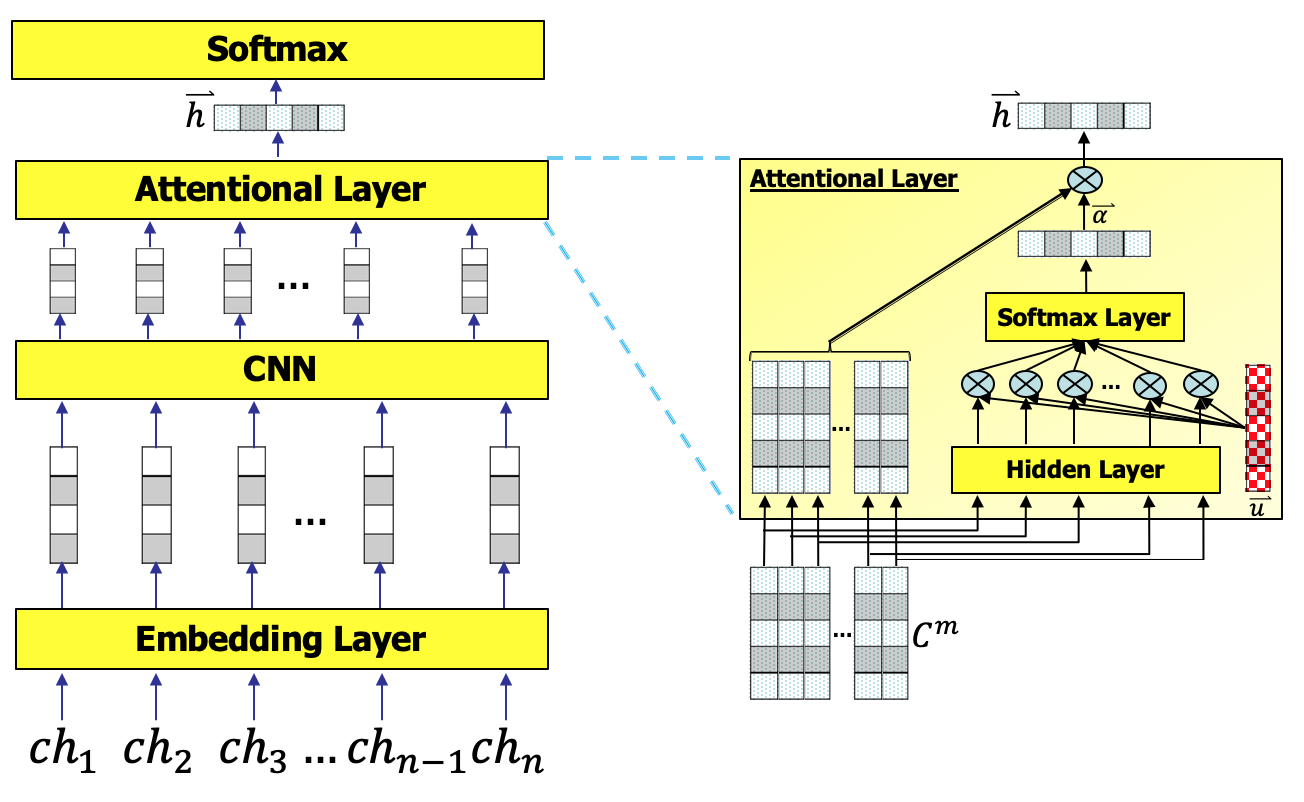}
    \caption{Our attentional-based CNN model.}
	\label{fig:attcnn}
\end{subfigure}
\caption{Neural network classifier architectures.}
\label{fig:baselines}
\end{figure*}

\section{Proposed Model} \label{ourwork}
Since many languages have unclear word boundaries, character n-grams, rather than words, have become widely used as input in LID systems \cite{lui2012langid,kocmi2017lanidenn,tromp2011graph,jurgens2017incorporating}. With this in mind, the LID problem can be defined as such: given a tweet $\mathit{tw}$ consisting of $n$ ordered characters ($\mathit{tw}=[ch_1, ch_2, ..., ch_n]$) selected within the vocabulary set $\mathit{char}$ of $V$ unique characters ($\mathit{char}=\{ch_1,ch_2, ..., ch_V\}$) and a set $\mathit{l}$ of $L$ languages ($\mathit{l}=\{l_1,l_2, ..., l_L\}$) , the aim is to predict the language $\mathit{\hat{l}}$ present in $tw$ using a classifier: 
\begin{equation}
\mathit{\hat{l}}=argmax_{l_i \in \mathit{l}}Score(l_i|tw),
\label{eqscore}
\end{equation}
\noindent where $Score(l_i|tw)$ is a scoring function quantifying how likely language $l_i$ was used given the observed message $tw$.

Most statistical LID systems follow the model of \cite{lui2012langid}. They start off by using what is called a \textit{one-hot encoding technique}, which represents each character $ch_i$ as a one-hot vector $\mathbf{x}_i^{oh} \in \mathbb{Z}_2^V$ according to the index of this character in $\mathit{char}$. This transforms $tw$ into a matrix $\mathbf{X}^{oh}$: 
\begin{equation}
\begin{split}
&\mathbf{X}^{oh}=[\mathbf{x}_1^{oh}, \mathbf{x}_2^{oh}, ..., \mathbf{x}_n^{oh} ] \in \mathbb{Z}_2^{V \times n}\\
&\text{where }\ \mathbf{x}_i{^{oh}}[j] = \begin{cases} 1, & \text{if } ch_i=\mathit{char}_j\\
0, & \text{otherwise}
\end{cases}
\end{split}
\end{equation}

\noindent The vector $\mathbf{X}^{oh}$ is passed to a feature extraction function, for example row-wise sum or tf-idf weighting, to obtain a feature vector $\mathbf{h}$. $\mathbf{h}$ is finally fed to a classifier model for either discriminative scoring (e.g. Support Vector Machine) or generative scoring (e.g. Naïve Bayes).\\

Unlike statistical methods, a typical neural network LID system, as illustrated in Figure \ref{fig:nnbaseline}, first pass this input through an embedding layer to map each character $ch_i \in tw$ to a low-dense vector $\mathbf{x}_i \in \mathbb{R}^d$, where $d$ denotes the dimension of character embedding. Given an input tweet $tw$, after passing through the embedding layer, we obtain an embedded matrix:
\begin{equation}
\mathbf{X}=[\ \mathbf{x}_1\ \mathbf{x}_2\ ... \mathbf{x}_n\ ] \in \mathbb{R}^{d \times n},
\end{equation}
\noindent The embedded matrix $\mathbf{X}$ is then fed through a neural network architecture, which transforms it into an output vector $\mathbf{h}=f(\mathbf{X})$ of length L that represents the likelihood of each language, and which is passed through a $\mathit{Softmax}$ function. This updates equation \ref{eqscore} as:
\begin{equation}
\begin{split}
\mathit{\hat{l}} 
&  = argmax_{l_i \in \mathit{l}}\mathit{Softmax}_i(\mathbf{h}).
\label{eqscore2}
\end{split}
\end{equation}

Tweets in particular are noisy messages which can contain a mix of multiple languages. To deal with this challenge, most previous neural network LID systems used deep sequence neural layers, such as an encoder-decoder \cite{jurgens2017incorporating} or a GRU \cite{kocmi2017lanidenn}, to extract global representations at a high computational cost. By contrast, we propose to employ a shallow (single-layer) convolution neural network (CNN) to locally learn region-based features. In addition, we propose to use an attention mechanism to proportionally merge together these local features for an entire tweet $tw$. We hypothesize that the attention mechanism will effectively capture which local features of a particular language are the dominant features of the tweet. There are two major advantages of our proposed architecture: first, the use of the CNN, which has the least number of parameters among other neural networks, simplifies the neural network model and decreases the inference latency; and second, the use of the attention mechanism makes it possible to model the mix of languages while maintaining a competitive performance. 


\subsection{ngam-regional CNN Model}\label{cnnmodel}

To begin, we present a traditional CNN with an ngam-regional constrain as our baseline. CNNs have been widely used in both image processing \cite{Lecun98} and NLP \cite{Collobert11}. The convolution operation of a filter with a region size $m$ is parameterized by a weight matrix $\mathbf{W}_{cnn} \in \mathbb{R}^{d_{cnn}\times md}$ and a bias vector $\mathbf{b}_{cnn} \in \mathbb{R}^{d_{cnn}}$, where $d_{cnn}$ is the dimension of the CNN. The inputs are a sequence of $m$ consecutive input columns in $\mathbf{X}$, represented by a concatenated vector $\mathbf{X}[i:i+m-1] \in \mathbb{R}^{md}$. The region-based feature vector $\mathbf{c}_i$ is computed as follows:
\begin{equation}
\begin{split}
& \mathbf{X}[i:i+m-1] = \mathbf{x}_i \oplus ... \oplus \mathbf{x}_{i+m-1},\\
& \mathbf{c}_i = g(\mathbf{W}_{cnn}\cdot\mathbf{X}[i:i+m-1]+\mathbf{b}_{cnn}),\\
\end{split}
\end{equation}
\noindent where $\oplus$ denotes a concatenation operation and $g$ is a non-linear function. The region filter is slid from the beginning to the end of $\mathbf{X}$ to obtain a convolution matrix $\mathbf{C}$:
\begin{equation}
{\mathbf{C}=[\mathbf{c}_1,...,\mathbf{c}_{n-m+1}] \in \mathbb{R}^{d_{cnn}\times (n-m+1)}.}\\
\label{C_matrix}
\end{equation}
\noindent The first novelty of our CNN is that we add a zero-padding constrain at both sides of $\mathbf{X}$ to ensure that the number of columns in $\mathbf{C}$ is equal to the number of columns in $\mathbf{X}$. Consequently, each $\mathbf{c}_i$ feature vector corresponds to an $\mathbf{x}_i$ input vector at the same index position $i$, and is learned from concatenating the surrounding $m$-gram embeddings. Particularly:
\begin{equation}
\begin{split}
  & 2p + (n-m+1) = n \\
  & p = \frac{m-1}{2}, \\
\end{split}
\end{equation}
where $p$ is the number of zero-padding columns. Finally, in a normal CNN, a row-wise max-pooling function is applied on $\mathbf{C}$ to extract the $d_{cnn}$ most salient features, as shown in Equation \ref{cnn_pool}. However, one weakness of this approach is that it extracts the most salient features out of sequence.
\begin{equation}
\begin{split}
\mathbf{h} & = f_{\mathit{CNN}}(\mathbf{X}) \\
& = \mathbf{h}_{cnn} \\
& = pooling_{max}(\mathbf{C}) \in  \mathbb{R}^{d_{cnn}}\\
\end{split}
\label{cnn_pool}
\end{equation}

\subsection{Attention Mechanism}\label{attcnn}
Instead of the traditional pooling function of Equation \ref{cnn_pool}, a second important innovation of our CNN model is to use an attention mechanism to model the interaction between region-based features from the beginning to the end of an input. Figure \ref{fig:attcnn} illustrates our proposed model. Given a sequence of regional feature vectors $\mathbf{C}=[\mathbf{c}_1,\mathbf{c}_2,...,\mathbf{c}_n]$ as computed in Equation \ref{C_matrix}, we pass it through a fully-connected hidden layer to learn a sequence of regional hidden vectors $\mathbf{H}=[\mathbf{h}_1,\mathbf{h}_2,...,\mathbf{h}_n] \in \mathbb{R}^{d_{hd} \times n}$ using Equation \ref{regional_hidden}.

\begin{equation}
{\mathbf{h}_i=g_2(\mathbf{W}_{hd} \cdot \mathbf{c}_i + \mathbf{b}_{hd}) \in \mathbb{R}^{d_{hd}},}\\
\label{regional_hidden}
\end{equation}
\noindent where $g_2$ is a non-linear activation function, $\mathbf{W}_{hd}$ and $\mathbf{b}_{hd}$ denote model parameters, and $d_{hd}$ is the dimension of the hidden layer. We followed Yang et al. \cite{Yang16} in employing a regional context vector $\mathbf{u} \in \mathbb{R}^{d_{hd}}$ to measure the importance of each window-based hidden vector. The regional importance factors are computed by:
\begin{equation}
{\mathbf{t}=\mathbf{H} \times \mathbf{u} \in \mathbb{R}^{{n}}.}\\
\end{equation}
\noindent The importance factors are then fed to a $\mathit{Softmax}$ layer to obtain the normalized weight:
\begin{equation}
{\alpha_i=\frac{\mathbf{t}_i}{\sum_{i'}\mathbf{t}_{i'}}.}\\
\end{equation}
The final representation of a given input is computed by a weighted sum of its regional feature vectors:
\begin{equation}
{\mathbf{h}=\sum_i \alpha_i\mathbf{c}_i}\\
\end{equation}

\section{Experimental Results} \label{results}

\begin{table}[]
\caption{Parameter settings}
\label{tab:para}
\centering
\begin{tabular}{lcc}
    \hline
    Parameter & our CNN & our att CNN \\ 
    \hline
     $d$ & $50$ & $50$ \\
     $g$ & $relu$ & $relu$ \\
     $g_2$ & $n.a.$ & $relu$ \\
     $d_{cnn}$ & $100$ & $100$ \\
     $m$ & $5$ & $5$ \\
     $p$ & $2$ & $2$ \\
     $d_{hd}$ & $n.a.$ & $100$ \\
     $lr$ & $0.001$ & $0.001$ \\
     $decay\_rate$ & $0.05$ & $0.05$ \\
     $max\_epochs$ & $512$ & $512$ \\
     $patience$ & $64$ & $64$ \\
     $clip\_rate$ & $5$ & $5$ \\ \hline
\end{tabular}
\end{table}

\begin{table*}
\caption{Benchmarking results.}
\small
\tabcolsep=0.1cm
\resizebox{\textwidth}{!}{\begin{tabular}{lccccr|ccccr|ccccr}
\hline
 \multirow{2}{*}{Model} & \multicolumn{5}{c|}{Small-scale dataset} & \multicolumn{5}{c|}{Medium-scale dataset} & \multicolumn{5}{c}{Large-scale dataset} \\ 
 & Acc & P & R & F1 & Speed & Acc & P & R & F1 & Speed & Acc & P & R & F1 & Speed\\ \hline
langid.py & 0.9229 & 0.9290 & 0.9229 & 0.9240 & 3710.96 & 0.9449 & 0.9475 & 0.9449 & 0.9454 & 3797.34 & 0.9483 & 0.9502 & 0.9483 & 0.9486 & 4630.59 \\ 
CLD2 & 0.8670 & 0.9624 & 0.8670 & 0.8997 & \textbf{43308.31} & 0.8784 & 0.9638 & 0.8784 & 0.9067 & \textbf{40287.01} & 0.8711 & 0.9527 & 0.8711 & 0.8952 & \textbf{42297.95}\\ 
CLD3 & 0.7284 & 0.8686 & 0.7284 & 0.7456 & 5911.94 & 0.7131 & 0.8792 & 0.7131 & 0.7333 & 6265.21 & 0.6976 & 0.9133 &  0.6976 & 0.7326 & 6139.86\\ 
LanideNN &  0.9304 & 0.9052 & 0.8984 & 0.9003 & 11.47 & 0.9414 & 0.9064 & 0.9005 & 0.9021 & 9.38 & 0.9370 & 0.8811 & 0.8745 & 0.8763 & 7.08 \\ 
EquiLID & 0.9244 & 0.9516 & 0.9244 & 0.9325 & 7.53 & 0.9430 & 0.9616 & 0.9430 & 0.9484 & 7.64 & 0.9489 & 0.9648 & 0.9489 & 0.9532 & 7.05 \\ 
langid.py (retrained) & 0.9013 & 0.9030 & 0.9013 & 0.9018 & 460.00 & 0.7866 & 0.8068 & 0.7866 & 0.7918 & 459.25 & 0.6731 & 0.7990 & 0.6731 & 0.6918 &  459.60\\ 
CNN & 0.9675 & 0.9677 & 0.9675 & 0.9675 & 39562.69 & 0.9832 & 0.9834 & 0.9832 & 0.9832 & 38427.42 & 0.9866 & 0.9867 &  0.9866 & 0.9866 & 31647.77\\ 
Attention CNN & \textbf{0.9712} & \textbf{0.9716} & \textbf{0.9712} & \textbf{0.9713} & 31782.44 & \textbf{0.9841} & \textbf{0.9842} & \textbf{0.9841} & \textbf{0.9842} & 24828.28 & \textbf{0.9915} & \textbf{0.9915} & \textbf{0.9915} & \textbf{0.9915} & 35266.82\\ \hline
\end{tabular}}
\label{table:ots final_results}
\end{table*}

\subsection{Benchmarks}\label{Benchmarking}

For the benchmarks, we selected five systems. We picked first the langid.py\footnote{\url{https://github.com/saffsd/langid.py}} library which is frequently used to compare systems in the literature. Since our work is in neural-network LID, we selected two neural network systems from the literature, specifically the encoder-decoder EquiLID\footnote{\url{https://github.com/davidjurgens/equilid}} system of \cite{jurgens2017incorporating} and the GRU neural network LanideNN\footnote{\url{https://github.com/tomkocmi/LanideNN}} system of \cite{kocmi2017lanidenn}. Finally, we included CLD2\footnote{\url{https://github.com/CLD2Owners/cld2}} and CLD3\footnote{\url{https://github.com/google/cld3}}, two implementations of the Naïve Bayes LID software used by Google in their Chrome web browser  \cite{lui2014accurate,jauhiainen2018automatic,bergsma2012language} and sometimes used as a comparison system in the LID literature \cite{blodgett2017dataset,jurgens2017incorporating,bergsma2012language,lui2012langid,kocmi2017lanidenn}. We obtained publicly-available implementations of each of these algorithms, and test them all against our three datasets. In Table \ref{table:ots final_results}, we report each algorithm's accuracy and F1 score, the two metrics usually reported in the LID literature. We also included precision and recall values, which are necessary for computing F1 score. And finally we included the speed in number of messages handled per second. This metric is not often discussed in the LID literature, but is of particular importance when dealing with a massive dataset such as ours or a massive streaming source such as Twitter.

We compare these benchmarks to our two models: the improved CNN as described in Section \ref{cnnmodel} and our proposed CNN model with an attention mechanism of Section \ref{attcnn}. These are labelled \textit{CNN} and \textit{Attention CNN} in Table \ref{table:ots final_results}. In both models, we filter out characters that appear less than $5$ times and apply a dropout approach with a dropout rate of $0.5$. ADAM optimization algorithm and early stopping techniques are employed during training. The full list of parameters and settings is given in Table \ref{tab:para}. It is worth noting that we randomly select this configuration without any tuning process.

\subsection{Analysis}

The first thing that appears from these results is the speed difference between algorithms. CLD3 and langid.py both can process several thousands of messages per second, and CLD2 is even an order of magnitude better, but the two neural network software have considerably worse performances, at less than a dozen messages per second. This is the efficiency trade-off of neural-network LID systems we mentioned in Section \ref{intro}; although to be fair, we should also point out that those two systems are research prototypes and thus may not have been fully optimized. 

In terms of accuracy and F1 score, langid.py, LanideNN, and EquiLID have very similar performances. All three consistently score above 0.90, and each achieves the best accuracy or the best F1 score at some point, if only by 0.002. By contrast, CLD2 and CLD3 have weaker performances; significantly so in the case of CLD3. In all cases, using our small-, medium-, or large-scale test set does not significantly affect the results.

All the benchmark systems were tested using the pre-trained models they come with. For comparison purposes, we retrained langid.py from scratch using the training and validation portion of our datasets, and ran the tests again. Surprisingly, we find that the results are worse for all metrics compared to using their pre-trained model, and moreover that using the medium- and large-scale datasets give significantly worse results than using the small-scale dataset. This may be a result of the fact the corpus the langid.py software was trained with and optimized for originally is drastically different from ours: it is a imbalanced dataset of 18,269 tweets in 9 languages. Our larger corpora, being more drastically different from the original, give increasingly worse performances. This observation may also explain the almost 10\% variation in performance of langid.py reported in the literature and reproduced in Table \ref{table:background}. The fact that the message handling performance of the library drops massively compared to its pre-trained results further indicates how the software was optimized to use its corpus. Based on this initial result, we decided not to retrain the other benchmark systems.

The last two lines of Table \ref{table:ots final_results} report the results of our basic CNN and our attention CNN LID systems. It can be seen that both of them outperform the benchmark systems in accuracy, precision, recall, and F1 score in all experiments. Moreover, the attention CNN outperforms the basic CNN in every metric (we will explore the benefit of the attention mechanism in the next subsection). In terms of processing speed, only the CLD2 system surpasses ours, but it does so at the cost of a 10\% drop in accuracy and F1 score. Looking at the choice of datasets, it can be seen that training with the large-scale dataset leads to a nearly 1\% improvement compared to the medium-sized dataset, which also gives a 1\% improvement compared to the small-scale dataset. While it is expected that using more training data will lead to a better system and better results, the small improvement indicates that even our small-scale dataset has sufficient messages to allow the network training to converge. 

\begin{table*}[ht] 
\caption{Tweets misclassified by the CNN but recognized by the Attention CNN} 
\label{tab:att_effect}
    \begin{tabular}{lccc} 
        \hline
        IDs & Tweets with attention values & att. CNN &  CNN\\ \hline
        $tw_{en_1}$ & \parbox[c]{.7\linewidth}{\includegraphics[width=\linewidth]{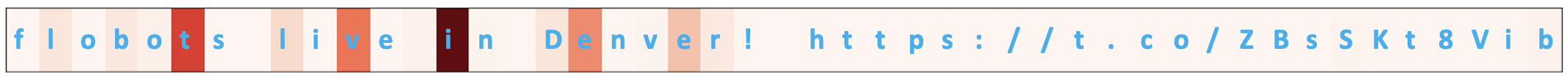}} & en & de\\ 
        $tw_{en_2}$ & \parbox[c]{.7\linewidth}{\includegraphics[width=\linewidth]{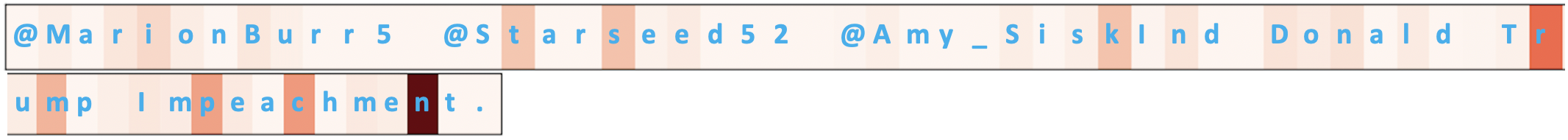}} & en & de\\ 
        $tw_{en_3}$ & \parbox[c]{.7\linewidth}{\includegraphics[width=\linewidth]{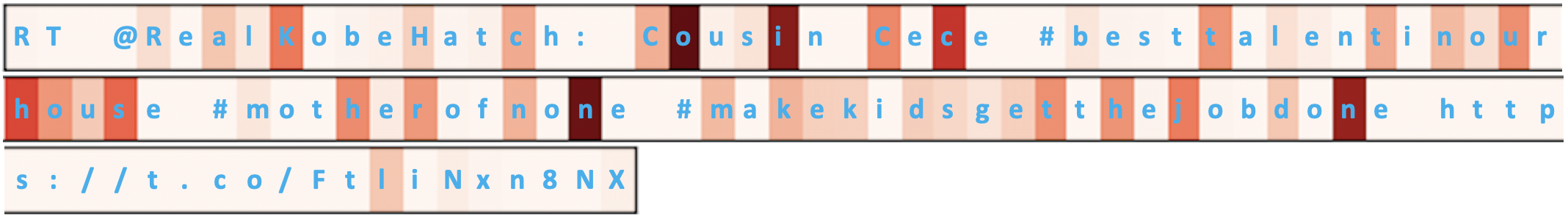}} & en & fr\\
        \hline
        $tw_{fr_1}$ & \parbox[c]{.7\linewidth}{\includegraphics[width=.8\linewidth]{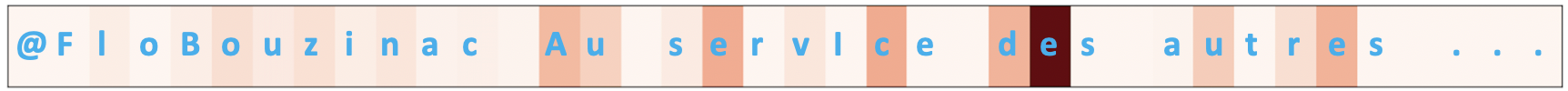}} & fr & de\\ 
        $tw_{fr_2}$ & \parbox[c]{.7\linewidth}{\includegraphics[width=\linewidth]{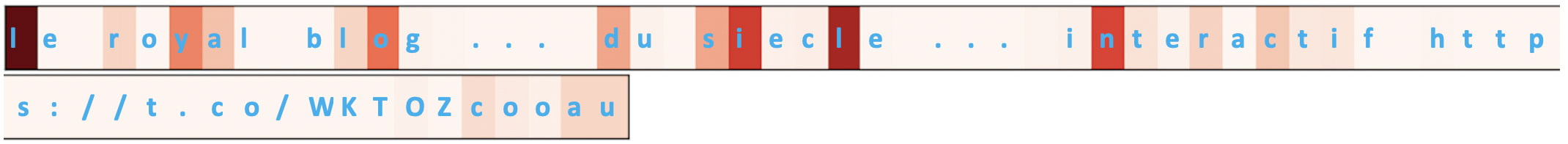}} & fr & ro\\ 
        $tw_{fr_3}$ & \parbox[c]{.7\linewidth}{\includegraphics[width=\linewidth]{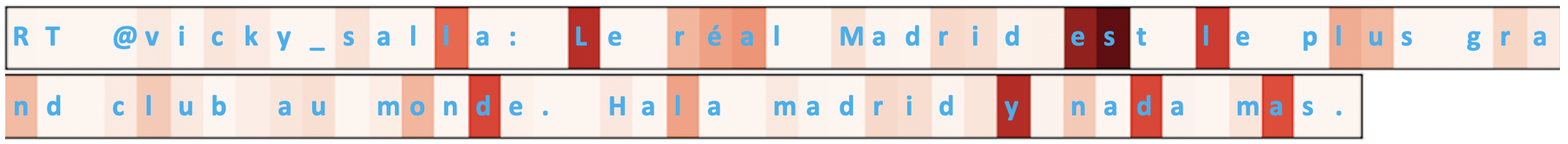}} & fr & es\\
        \hline
        $tw_{vi_1}$ & \parbox[c]{.7\linewidth}{\includegraphics[width=.5\linewidth]{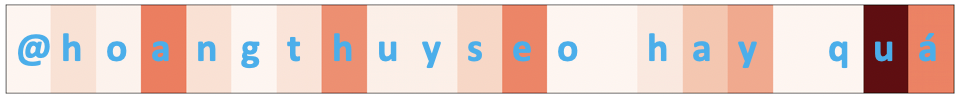}} & vi & es \\ 
        $tw_{vi_2}$ & \parbox[c]{.7\linewidth}{\includegraphics[width=\linewidth]{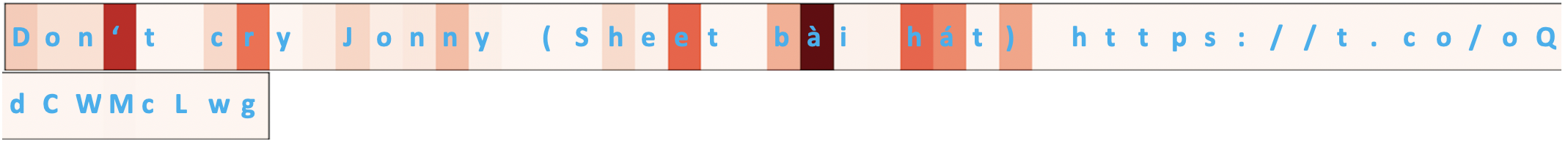}} & vi & hu \\ 
        $tw_{vi_3}$ & \parbox[c]{.7\linewidth}{\includegraphics[width=\linewidth]{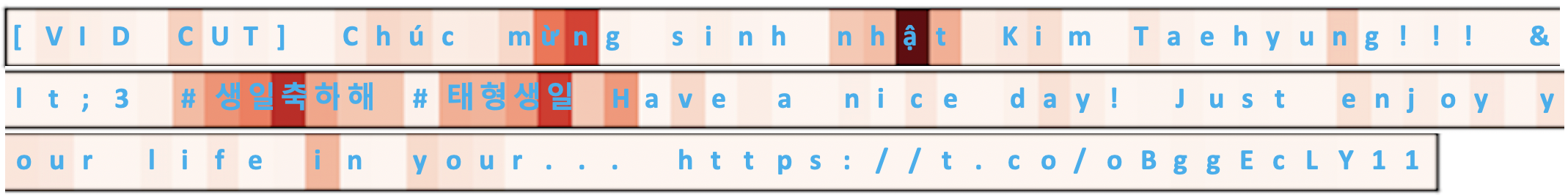}} & vi & ko\\
        \hline
    \end{tabular}
\end{table*}

\subsection{Impact of Attention Mechanism}
We can further illustrate the impact of our attention mechanism by displaying the importance factor $\alpha_i$ corresponding to each character $ch_i$ in selected tweets. Table \ref{tab:att_effect} shows a set of tweets that were correctly identified by the attention CNN but misclassified by the regular CNN in three different languages: English, French, and Vietnamese. The color intensity of a letter's cell is proportional to the attention mechanism's normalized weight $\alpha_i$, or on the focus the network puts on that character. In order words, the attention CNN puts more importance on the features that have the darkest color. 

The case studies of Table \ref{tab:att_effect} show the noise-tolerance that comes from the attention mechanism. It can be seen that the system puts virtually no weight on URL links (e.g. $tw_{en_1}$, $tw_{fr_2}$, $tw_{vi_2}$), on hashtags (e.g. $tw_{en_3}$), or on usernames (e.g. $tw_{en_2}$, $tw_{fr_1}$, $tw_{vi_1}$). We should emphasize that our system does not implement any text preprocessing steps; the input tweets are kept as-is. Despite that, the network learned to distinguish between words and non-words, and to focus mainly on the former. In fact, when the network does put attention on these elements, it is when they appear to use real words (e.g. ``star" and ``seed"  in the username of $tw_{en_2}$, ``mother" and ``none" in the hashtag of $tw_{en_3}$). This also illustrates how the attention mechanism can pick out fine-grained features within noisy text: in those examples, it was able to focus on real-word components of longer non-word strings.


The examples of Table \ref{tab:att_effect} also show that the attention CNN learns to focus on common words to recognize languages. Some of the highest-weighted characters in the example tweets are found in common determiners, adverbs, and verbs of each language. These include ``\textbf{i}n"~($tw_{en_1}$), ``d\textbf{e}s"~($tw_{fr_1}$), ``\textbf{l}e"~($tw_{fr_2}$), ``\textbf{es}t"~($tw_{fr_3}$),  \VI{``q\textbf{u}á"~($tw_{vi_2}$)}, and \VI{``nh\textbf{ấ}t"~($tw_{vi_3}$)}. These letters and words significantly contribute in identifying the language of a given input.  

Finally, when multiple languages are found within a tweet, the network successfully captures all of them. For example, $tw_{fr_3}$ switches from French to Spanish and $tw_{vi_2}$ mixes both English and Vietnamese. In both cases, the network identifies features of both languages; it focuses strongly on ``\textbf{es}t" and ``\textbf{y}" in $tw_{fr_3}$, and on ``Don\textbf{'}t" and \VI{``b\textbf{à}i"} in $tw_{vi_2}$. The message of $tw_{vi_3}$ mixes three languages, Vietnamese, English, and Korean, and the network focuses on all three parts, by picking out \VI{``nh\textbf{ậ}t"} and \VI{``m\textbf{ừ}ng"} in Vietnamese, ``\#생일\textbf{축}하해" and ``\#태형생\textbf{일}" in Korean, and ``$\textbf{h}ave$" in English. Since our system is setup to classify each tweet into a single language, the strongest feature of each tweet wins out and the message is classified in the corresponding language. Nonetheless, it is significant to see that features of all languages present in the tweet are picked out, and a future version of our system could successfully decompose the tweets into portions of each language. 

\section{Conclusion} \label{conclusion}
In this paper, we first demonstrated how to build balanced, automatically-labelled, and massive LID datasets. These datasets are taken from Twitter, and are thus composed of real-world and noisy messages. We applied our technique to build three datasets ranging from hundreds of thousands to tens of millions of short texts. Next, we proposed our new neural LID system, a CNN-based network with an attention mechanism to mitigate the performance bottleneck issue while still maintaining a state-of-the-art performance. The results obtained by our system surpassed five benchmark LID systems by 5\% to 10\%. Moreover, our analysis of the attention mechanism shed some light on the inner workings of the typically-black-box neural network, and demonstrated how it helps pick out the most important linguistic features while ignoring noise. All of our datasets and source code are publicly available at \url{https://github.com/duytinvo/LID\_NN}.

\newpage

\bibliography{emnlp2018}
\bibliographystyle{icml2019}

\end{document}